%% file: main.tex
\documentclass[10pt,twocolumn,letterpaper]{article}

\usepackage{cvpr}
\usepackage{times}
\usepackage{epsfig}
\usepackage{graphicx}
\usepackage{amsmath}
\usepackage{amssymb}
\usepackage{amscd}

\usepackage{url}

\usepackage{microtype}
\usepackage{graphicx}
\usepackage{subfigure}
\usepackage{booktabs} 
\usepackage{algorithm}
\usepackage{algorithmic}
\usepackage{multirow}
\usepackage{xcolor}
\usepackage{amsmath}
\usepackage{mathtools}
\usepackage{tabularx}
\newcolumntype{C}{>{\centering\arraybackslash}X}
\usepackage{enumitem}
\usepackage{bm}
\usepackage[utf8]{inputenc} 
\usepackage[T1]{fontenc}    
\usepackage{booktabs}       
\usepackage{amsfonts}       
\usepackage{nicefrac}       
\usepackage{microtype}      
\usepackage{amsthm}
\usepackage{xr}

\DeclarePairedDelimiter\norm{\lVert}{\rVert}

\newtheorem{theorem}{Theorem}

\usepackage{authblk}

\usepackage[breaklinks=true,bookmarks=false]{hyperref}

\cvprfinalcopy 

\def\cvprPaperID{****} 

\begin{document}

\title{AE-OT-GAN: Training GANs from data specific latent distribution}

\author[1]{Dongsheng An}
\author[1]{Yang Guo}
\author[2]{Min Zhang}
\author[1]{Xin Qi}
\author[3]{Na Lei}
\author[4]{Shing-Tung Yau}
\author[1]{Xianfeng Gu}
\affil[1]{Department of Computer Science, Stony Brook University}
\affil[2]{Brigham and Women's Hospital, Harvard Medical School}
\affil[3]{DLUT-RU, Dalian University of Technology}
\affil[4]{Department of Mathematics, Harvard University}
 
\maketitle
\vspace{-5mm}
\begin{abstract}
   Though generative adversarial networks (GANs) are prominent models to generate realistic and crisp images, they are unstable to train and suffer from the mode collapse/mixture. The problems of GANs come from approximating the intrinsic discontinuous distribution transform map with continuous DNNs. 
  The recently proposed AE-OT model addresses the discontinuity problem by explicitly computing the discontinuous optimal transform map in the latent space of the autoencoder. Though have no mode collapse/mixture, the generated images by AE-OT are blurry. In this paper, we propose the AE-OT-GAN model to utilize the advantages of the both models: generate high quality images and at the same time overcome the mode collapse/mixture problems. Specifically, we firstly embed the low dimensional image manifold into the latent space by training an autoencoder (AE). Then the extended semi-discrete optimal transport (SDOT) map from the uniform distribution to the empirical latent distribution is used to generate new latent codes. Finally, our GAN model is trained to generate high quality images from the latent distribution induced by the extended SDOT map. The distribution transform map from this dataset related latent distribution to the data distribution will be continuous, and thus can be well approximated by the continuous DNNs. Additionally, the paired data between the latent codes and the real images gives us further restriction about the generator and stabilizes the training process. Experiments on simple MNIST dataset and complex datasets like CIFAR10 and CelebA show the advantages of the proposed method.
\end{abstract}

\input{introduction.tex}
\input{related_work.tex}
\input{algorithm.tex}

\input{experiments.tex}

\input{conclusion.tex}

{\small
\bibliographystyle{ieee_fullname}
\bibliography{egbib}
}

\end{document}

%% file: introduction.tex
\section{Introduction}
\label{sec:intro}
\vspace{-3mm}
Image generation has been one of the core topics in the area of computer vision for a long time. Thanks to the quick development of deep learning, numerous generative models are proposed, including encoder-decoder based models \cite{kingma2013vae, Tolstikhin2018WAE, An2019AEOT}, generative adversarial networks (GANs) \cite{Ian2014GAN, infogan2016alec, Xiao2018Bourgan, radford2015DCGAN, arjovsky2017wgan, gulrajani2017improvedgan}, density estimator based models \cite{Oord2016PixelCNN, nice2014, nvp2017, glow2018} and energy based models \cite{lecun2006energy, zhu1998frame, xie2016dgn, ebm2019}. The encoder-decoder based models and GANs are the most prominent ones due to their capability to generate high quality images.


\begin{figure}[t]
    \centering
    \includegraphics[width=0.8\linewidth]{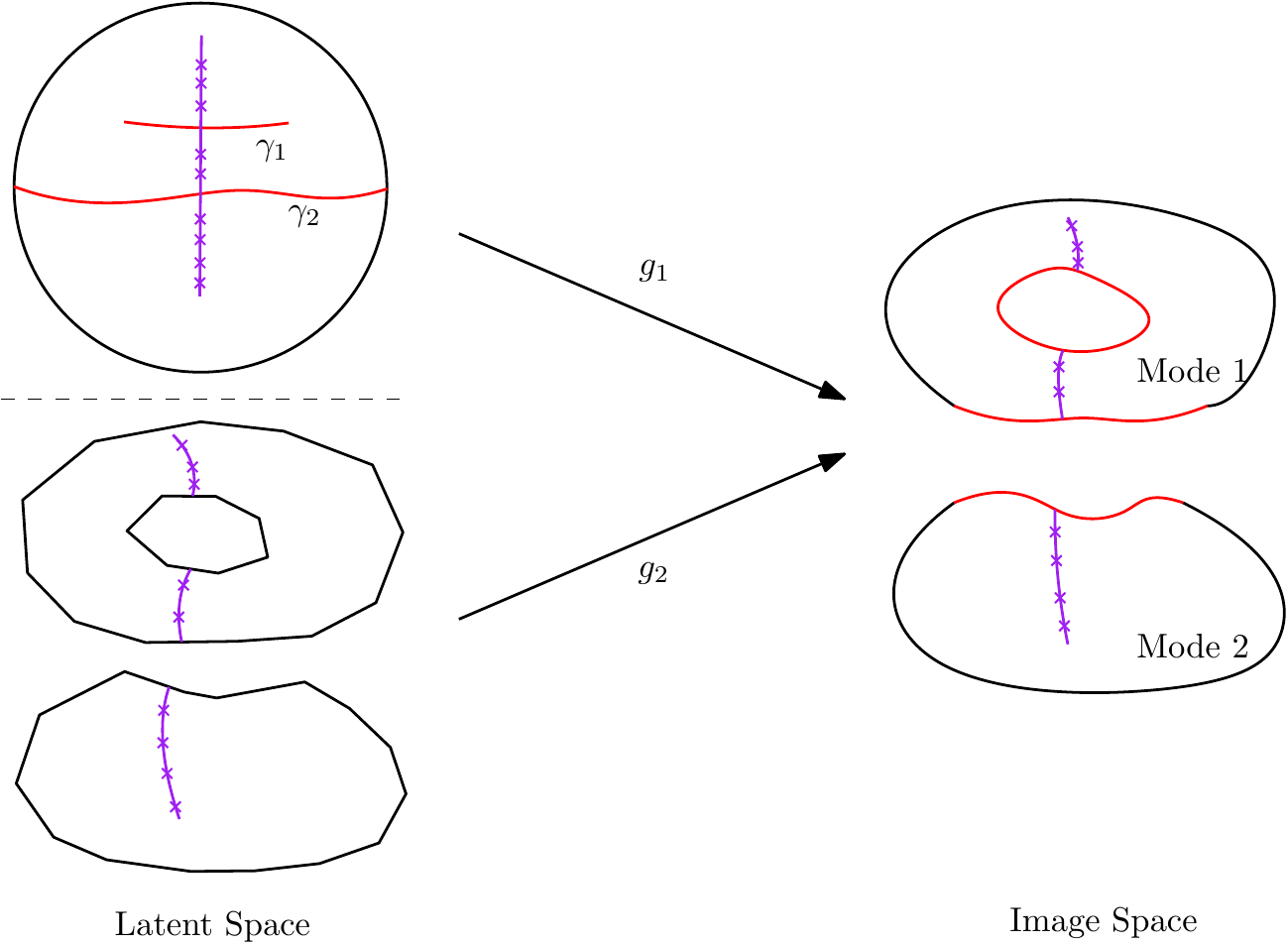}
    \caption{Distribution transport maps from two different latent distributions to the same data distribution. $g_1$ maps a \textit{unimodal} latent distribution on the left to the data distribution on the right. Difference in the topology of their supporting manifolds will cause discontinuities of the map, which is hard to approximate by continuous neural networks. The singular set of $g_1$ consists of $\gamma_1$ and $\gamma_2$ (shown in red): continuous samplings of the source distribution are mapped to three disjoint segments (shown in purple). On the other hand, $g_2$ samples from a \textit{suitably supported} latent distribution and is less likely to suffer from the discontinuity problem. Thus it can be well approximated by neural networks.}
    \label{fig:regularity}
\vspace{-5mm}
\end{figure}

Intrinsically, the generator in a generative model aims to learn the real data distribution supported on the data manifold \cite{Tenenbaum2000}. Suppose the distribution of a specific class of natural data $\nu_{gt}$ is concentrated on a low dimensional manifold $\chi$ embedded in the high dimensional data space. The encoder-decoder methods first attempt to embed the data into the latent space $\Omega$ through the encoder $f_\theta$, then samples from the latent distribution are mapped back to the manifold to generate new data by decoder $g_\xi$. While GANs, which have no encoder, directly learn a map (generator) that transports a given prior low dimensional distribution to $\nu_{gt}$. 

Usually, GANs are unstable to train and suffer from mode collapse \cite{goodfellow2016nips, lucic2018gans}. The difficulties come from the fact that the generator of a GAN model is trained to approximate the discontinuous distribution transport map from the {\it unimodal Gaussian distribution} to the {\it real data distribution} by the continuous neural networks \cite{Xiao2018Bourgan, An2019AEOT, Mahyar2018discontinuouity}. In fact, when the supporting manifolds of the source and target distributions differ in topology or convexity, the OT map between them will be discontinuous \cite{villani2008optimal}, as illustrated in the map $g_1$ of Fig. \ref{fig:regularity}. In practice, distribution transport maps can have complicated singularities, even when the ambient dimension is low (see e.g. \cite{figalli2010regularity}). This poses a great challenge for the generator training in standard GAN models. 

To tackle the mode collapse and mode mixture problems caused by discontinuous transport maps, the authors of \cite{An2019AEOT} proposed the AE-OT model. In this model, an autoencoder is used to map the images manifold $\chi$ into the latent manifold $\Omega$. Then, the semi-discrete optimal transport (SDOT) map $T$ from the uniform distribution $Uni([0,1]^d)$ to the latent empirical distribution is explicitly computed via convex optimization approach. Then a piece-wise linear extension map of the SDOT, denoted by $\Tilde{T}$, pushes forward the uniform distribution to a continuous latent distribution $\mu$, which in turn gives a good approximation of the latent distribution $\mu_{gt}=f_{\theta\#}\nu_{gt}$ ($f_{\theta\#}$ means the push forward map induced by $f_\theta$). Composing the continuous decoder $g_\xi$ and discontinuous $\Tilde{T}$ together, i.e. $g_\xi \circ \Tilde{T}(w)$, where $w$ is sampled from uniform distribution, this model can generate new images. Though have no mode collapse/mixture, the generated images look blurry. The framework of AE-OT is shown as follows:
\[
\begin{CD}
(\nu_{gt},\chi) @>{f_\theta}>> (\mu_{gt}/\mu,\Omega) @>{g_\xi}>> (\nu_{gt}, \chi) \\
@. @A{\Tilde{T}}AA @.\\
@. (Uni([0,1]^d),~ [0,1]^d) @.
\end{CD}
\]

In this work we propose the AE-OT-GAN framework to combine the advantages of the both models and generate high quality images without mode collapse/mixture. Specifically, after the training of the autoencoder and the computation of the extended SDOT map, we can directly sample from the latent distribution $\mu$ by applying $\Tilde{T}(w)$ on the uniform distribution to train the GAN model. In contrast to the conventional GAN models, whose generators are trained to transport the latent Gaussian distribution to the data manifold distributions, our GAN model sample from the data inferred latent distribution $\mu$. The distribution transport map from $\mu$ to the data distribution $\nu_{gt}$ is continuous and thus can be well approximated by the generator (parameterized by CNNs), as shown in $g_2$ of Fig. \ref{fig:regularity}. Moreover, the decoder of the pre-trained autoencoder gives a warm start of the generator, so that the Kullback–Leibler divergence between real and fake batches of images have non-vanishing overlap in their supports during the training phase. Furthermore, the content loss and feature loss between paired latent codes and real input images regularize the adversarial loss and stabilize the GAN training. Experiments have shown efficacy and efficiency of our proposed model.  

The contributions of the current work can be summarized as follows:
\textbf{(1)} This paper proposes a novel AE-OT-GAN  model that combines the strengths of AE-OT model and GAN model. It eliminates the mode collapse/mixture of GAN and removes the blurriness of the images generated by AE-OT. 
\textbf{(2)} The decoder of the autoencoder provides a good initialization of the generator of GAN. The number of iterations required to reach the equilibrium has been reduced by more than 100 times compared to typical GANs.
\textbf{(3)} In addition to the adversarial loss, the explicit correspondence between the latent codes and the real images provide auxiliary constraints, namely the content loss, to the generator.
\textbf{(4)} Our experiments demonstrate that our model can generate images consistently better than or comparable to the results of state-of-the-art methods.

%% file: related_work.tex
\section{Related Work}
\begin{figure*}[ht]
\centering
    \includegraphics[width=.75\linewidth]{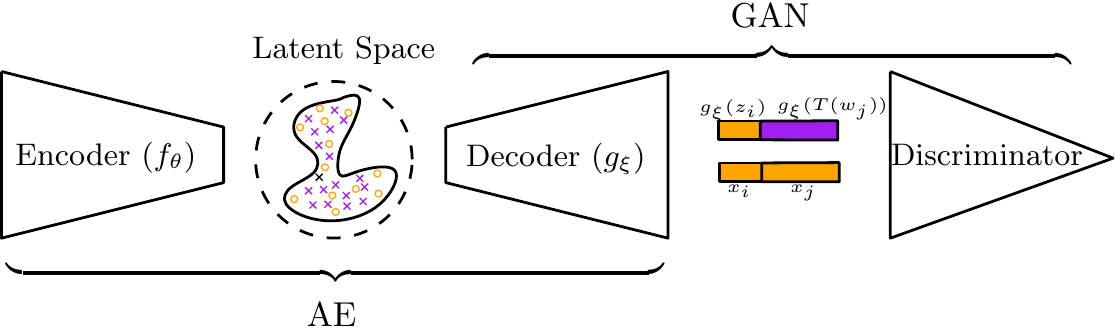}
    \caption{The framework of the proposed method. Firstly, the autoencoder is trained to embed the images into the latent space, the real latent codes are shown as the orange circles. Then we compute extended semi-discrete OT map $\Tilde{T}$ to generate new latent codes in the latent space (the purple crosses). Finally, our GAN model is trained from the latent distribution induced by $\Tilde{T}$ to the image distribution. Here the generator is just the decoder of the autoencoder. The fake batch (the bar with orange and purple colors) to train the discriminator is composed of two parts: the reconstructed images $g_\xi(z_i)$ of the real latent codes  and the generated images $g_\xi(\Tilde{T}(w))$ from the randomly generated latent codes with $w$ sampled from uniform distribution. The real batch (the bar with only orange color) is also composed of two parts: the real images $x_i$ corresponding to $z_i$, and the randomly selected images $x_j$.
    }
    \label{fig:framework}
    \vspace{-5mm}
\end{figure*}
\label{sec:related_work}
The proposed method in this paper is highly related to encoder-decoder based generation models, the generative adversarial networks (GANs), conditional GANs and the hybrid models that take the advantages of above. 

\textbf{Encoder-decoder architecture} A breakthrough for image generating comes from the scheme of Variational Autoencoders (VAEs) (e.g. \cite{kingma2013vae}), where the decoders approximate real data distributions from a Gaussian distribution in a variational approach (e.g \cite{kingma2013vae} and \cite{rezende2014stochastic}). Latter Yuri Burda et al. \cite{burda2015iwae} lower the requirement of latent distribution and propose the importance weighted autoencoder (IWAE) model through a different lower bound. Bin and David \cite{dai2018diagvae} propose that the latent distribution of VAE may not be Gaussian and improve it by firstly training the original model and then generating new latent code through the extended ancestral process. Another improvement of the VAE is the VQ-VAE model \cite{Oord2017vqvae}, which requires the encoder to output discrete latent codes by vector quantisation, then the posterior collapse of VAEs can be overcome. By multi-scale hierarchical organization, this idea is further used to generate high quality images in VQ-VAE-2 \cite{Ali2019vqvae2}. 
In \cite{Tolstikhin2018WAE}, the authors adopt the Wasserstein distance in the latent space to measure the distance between the distribution of the latent code and the given one and generate images with better quality.
Different from the the VAEs, the AE-OT model \cite{An2019AEOT} firstly embed the images into the latent space by autoencoder, then an extended semi-discrete OT map is computed to generate new latent code based on the fixed ones. Decoded by the decoder, new images can be generated. 
Although the encoder-decoder based methods are relatively simple to train, the generated images tend to be blurry.

\textbf{Generative adversarial networks} The GAN model \cite{Ian2014GAN} tries to alternatively update the generator, which maps the noise sampled from a given distribution to real images, and the discriminator differentiates the difference between the generated images and the real ones. If the generated images successfully fool the discriminator, we say the model is well trained. Later, \cite{radford2015DCGAN} proposes a deep convolutions neural network (DCGAN) to generate images with better quality. 
While being a powerful tool in generating realistic samples, GANs can be hard to train and suffer from mode collapse problem \cite{goodfellow2016nips}. After delicate analysis, \cite{arjovsky2017wgan} points out that it is the KL divergence the original GAN used causes these problems. Then the authors introduce the celebrated WGAN, which makes the whole framework easy to converge. To satisfy the lipschitz continuity required by WGAN, a lot of methods are proposed, including clipping \cite{arjovsky2017wgan}, gradient penalty \cite{gulrajani2017improvedgan}, spectral normalization \cite{miyato2018spectral} and so on. Later, Wu et al. \cite{wu2018wgan-div} use the wasserstein divergence objective, which get rid of the lipschitz approximation problem and get a better result. Instead $L_1$ cost adopted by WGAN, Liu et.al \cite{liu2019wganqc} propose the WGAN-QC by taking the $L_2$ cost into consideration. Though various GANs can generate sharp images, they will theoretically encounter the mode collapse or mode mixture problem \cite{goodfellow2016nips, An2019AEOT}.

\textbf{Hybrid models} To solve the blurry image problem of encoder-decoder architecture and the mode collapse/mixture problems of GANs, a natural idea is to compose them together. 
Larsen et al. \cite{Larsen2016VAEGAN} propose to combine the variational autoencoder with a generative adversarial network, and thus generate images better than VAEs. \cite{makhzani2015aae} matches the aggregated posterior of the hidden code vector of the autoencoder with an arbitrary prior distribution by a discriminator and then applies the model into tasks like semi-supervised classification and dimensionality reduction. 
 BiGAN \cite{Donahue2017bigan}, with the same architecture with ours, uses the discriminator to differentiate both the generated images and the generated latent code. Further, by utilizing the BigGAN generator \cite{Brock2019biggan}, the BigBiGAN \cite{Donahue2019bigbigan} extends this method to generate much better results. Here we also treat the BourGAN \cite{Xiao2018Bourgan} as a hybrid model, because it firstly embeds the images into latent space by Bourgain theorem, then trains the GAN model by sampling from the latent space using the GMM model.
 
Conditional GANs are another kind of hybrid models that can also be treated as image-to-image transformation. For example, using an encoder-decoder architecture to build the connection between paired images and then differentiating the decoded images with the real ones by a discriminator, \cite{isola2017image} is able to transform images of different styles. Further, SRGAN \cite{Ledig2017SUR} uses similar architecture to get super resolution images from their low resolution versions. The SRGAN model is the most similar work to ours, as it also utilizes the content loss and adversarial loss. The main differences between this model and ours including: (i) SRGAN just uses the paired data, while the proposed method use both the paired data and generated new latent code to train the model; (ii) the visually meaningful features used by SRGAN are extracted from the pre-trained VGG19 network \cite{Simonyan2014VGG}, while in our model, they come from the encoder itself. This makes them more reasonable especially under the scenes where the datasets are not included in those used to train the VGG.

%% file: algorithm.tex
\vspace{-1mm}
\section{The Proposed Method}
\vspace{-1mm}
\label{sec:alg}

In this section, we explain our proposed AE-OT-GAN model in detail. There are mainly three modules, an autoencoder (AE), an optimal transport mapper (OT) and a GAN model.
Firstly, an AE model is trained to embed the data manifold $\chi$ into the latent space. 
At the same time, the encoder $f_\theta$ pushes forward the ground-truth data distribution $\nu_{gt}$ supported on $\chi$ to the ground-truth latent distribution $\mu_{gt}$ supported on $\Omega$ in the latent space. Secondly, we compute the semi-discrete OT map from the uniform distribution to the empirical latent distribution. By extending the SDOT map, we can construct the continuous distribution $\mu$ that approximates the ground-truth latent distribution $\mu_{gt}$ well. Finally, starting from $\mu$ as the latent distribution, our GAN model is trained to generate both realistic and crisp images. The pipeline of our proposed model is illustrated in Fig. \ref{fig:framework}. In the following, we will explain the three modules one by one.

\subsection{Data Embedding with Autoencoder}
\label{subsec:mani_emb}
We model the real data distribution as a probability measure $\nu_{gt}$ supported on an $r$ dimensional manifold $\chi$ embedded in the $D$ dimensional Euclidean space $\mathbb{R}^{D}$ (ambient space) with $r\ll D$. 

In the first step of our AE-OT-GAN model, we train an autoencoder (AE) to embed the real data manifold $\chi$ to be the latent manifold $\Omega$. In particular, training the AE model is equivalent to compute the encoding map $f_\theta$ and decoding map $g_\xi$  
\[
\begin{CD}
(\nu_{gt}, \chi)@>{f_\theta}>> (\mu_{gt},\Omega) @>{g_\xi}>> (\nu_{gt}, \chi) 
\end{CD}
\]
by minimizing the loss function:
\[
    \mathcal{L}(\theta,\xi):= \sum_{i=1}^n \|x_i - g_\xi\circ f_\theta(x_i)\|^2,
\]
with $f_\theta$ and $g_\xi$ parameterized by standard CNNs ($\theta$ and $\xi$ are the parameters of the networks, respectively). Given densely sampling from the image manifold (detailed explanation is included in the supplementary) and ideal optimization (namely the loss function goes to $0$), 
$f_\theta\circ g_\xi$ coincides with the identity map. 
After training, $f_\theta$ is a continuous, convertible map, namely a \emph{homeomorphism}, and $g_\xi$ is the inverse homeomorphism. This means $f_\theta: \chi\to\Omega$ is an embedding, and pushes forward $\nu_{gt}$ to the latent data distribution $\mu_{gt}:=f_{\theta\#}\nu_{gt}$. In practice, we only have the empirical data distribution given by $\hat{\nu}_{gt}=\frac{1}{n}\sum_{i=1}^n \delta(x-x_i)$, which is push forward to be the empirical latent distribution  $\hat{\mu}_{gt}=\frac{1}{n}\sum_{i=1}^n \delta(z-z_i)$, where $n$ is the number of samples.

\subsection{Constructing $\mu$ with Semi-Discrete OT Map}
\label{subsec:extended}
\begin{figure}[t]
    \centering
    \includegraphics[width=\linewidth]{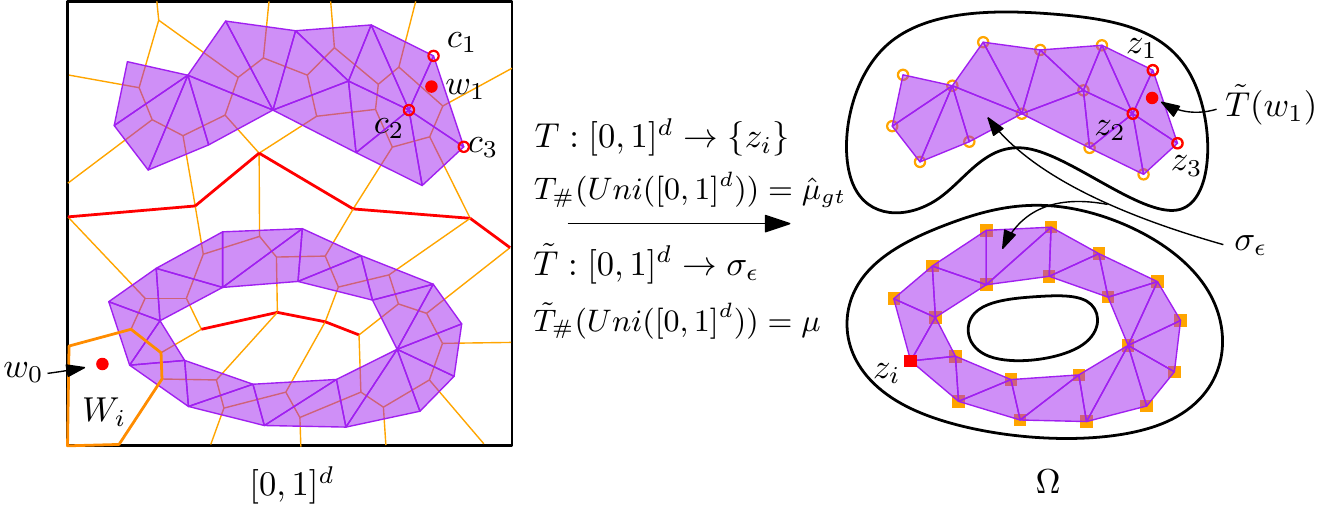}
    \caption{OT map $T$ and the extended OT map $\tilde{T}$ in 2D case. Here $T$ maps points (e.g. $w_0$) in each polyhedral cell $W_i$ (orange cells on the left) to the corresponding latent code $z_i$ (circles and squares on the right). The piece-wise linear $\tilde{T}$ maps triangulated regions in $[0,1]^d$ to the simplicial complex $\sigma_\varepsilon$ in the latent space (shown in purple). Given the barycenters $c_i$'s of each $W_i$'s, each triangle $\Delta c_ic_jc_k$ is mapped to the corresponding simplex $[z_i,z_j,z_k]$. For example, $w_1$ in the triangle $\Delta c_0c_1c_2$ is mapped to $\tilde{T}(w_1)$ in the simplex $[z_0,z_1,z_2]$. Red lines in $[0,1]^d$ illustrate the singular set of $T$, which corresponds to the pre-image of gaps or holes in $\Omega$.} 
    \label{fig:alg_rips}
    \vspace{-3mm}
\end{figure}

In this section, from the empirical latent distribution $\hat{\mu}_{gt}$, we construct a continuous latent distribution $\mu$ following \cite{An2019AEOT} such that
(i) it generalizes $\hat{\mu}_{gt}$ well, so that all of the modes are covered by the support of $\mu$ (ii) the support of $\mu$ has similar topology to that of $\mu_{gt}$, which ensures that the transport map from $\mu$ to $\nu_{gt}$ is continuous and (iii) it is efficient to sample from $\mu$. 

To obtain $\mu$, the semi-discrete OT map $T$ from the uniform distribution $Uni([0,1]^d)$ to $\hat{\mu}_{gt}$ is firstly computed. Here $d$ is the dimension of the latent space. By extending $T$ to be a piece-wise linear map $\tilde{T}$, we can construct $\mu$ as the push forward distribution of $Uni([0,1]^d)$ under $\tilde{T}$: 
\[
\begin{CD}
(Uni([0,1]^d), [0,1]^d)@>{\tilde{T}}>> (\mu,\Omega)
\end{CD}
\]


In the first step, we compute the semi-discrete OT map $T:[0,1]^d \rightarrow \Omega$, with $T_{\#}{Uni([0,1]^d)} =\hat{\mu}_{gt}$. Under $T$, the continuous domain of $[0,1]^d$ is decomposed into cells $\{W_i\}$ with $T(w)=z_i,~ \forall w \in W_i$, with the Lebesgue measure of each $W_i$ to be $\frac{1}{n}$. The cell structure is shown in the left frame of Fig. \ref{fig:alg_rips} (the orange cells). Computational details of $T$ can be found in the supplementary material and \cite{An2019AEOT}.

Secondly, we extend the image domain of $T$ from the discrete latent codes $\{z_i\}$ to a continuous neighborhood $\sigma_\varepsilon$, which serves as the supporting manifold of $\mu$.  Specifically, we construct a simplicial complex $\sigma_\varepsilon$ from the latent codes $\{z_i\}$. Here $\varepsilon > 0$ is a constant. The 0-skeleton of $\sigma_\varepsilon$, represented by $\sigma_\varepsilon^{(0)}$, is the set of all latent codes $\{z_i\}$. 
The we define its k-skeletons $\sigma_\varepsilon^{(k)}$ by $\sigma_\varepsilon^{(k)} = \{[z_1,z_2,\dots,z_k]\mid \norm{z_i - z_j}_2\leq \varepsilon, \forall 1\leq i < j \leq k\}$ for $0<k\leq d$. The right frame of Fig. \ref{fig:alg_rips} shows an example of $\sigma_\varepsilon$. By assuming that the latent code is densely sampled from the latent manifold $\Omega$ and with an appropriate $\varepsilon$, $\sigma_\varepsilon$ will have consistent "hole" and "gap" structure with $\Omega$, in the sense of homology equivalence. Details are described in the supplementary material. 

Finally, we define the piece-wise linear extended OT map $\tilde{T}:[0,1]^d\rightarrow \sigma_\varepsilon$. 
Given a random sample $w$ sampled from ${Uni}([0,1]^d)$, we can find the cell $W_i$ containing it. By computing the barycentric parameters $\lambda_j$'s with respect to the nearby mass centers $c_j$'s of the cells $W_j$'s, i.e. compute $\lambda_j$'s such that $w = \Sigma \lambda_j c_j$ with $0\leq \lambda_j\leq 1$ and $\Sigma \lambda_j = 1$. Here $W_j$ represents the neighbour of $W_i$. Then $w$ is mapped to $\tilde{T}(w) \coloneqq \Sigma \lambda_j T(c_j) = \Sigma \lambda_j z_j$ if the corresponding $z_j$'s form a simplex of $\sigma_\varepsilon$. Otherwise we map $w$ to $z_i$, i.e. $\tilde{T}(w)\coloneqq T(c_i) = z_i$. As illustrated in Fig. \ref{fig:alg_rips}, compared to the many-to-one semi-discrete OT map $T$, $\tilde{T}$ maps samples within the triangular areas (the purple triangles on the left frame) in $[0,1]^d$ \textit{linearly} to the corresponding simplices in $\sigma_\varepsilon$ (the purple triangles on the right frame) in a bijective manner. We denote the pushed forward distribution under $\tilde{T}$ as $\mu_\varepsilon \coloneqq \tilde{T}_\# {Uni}([0,1]^d)$. 
\begin{theorem}
The 2-Wasserstein distance between $\mu_\varepsilon$ and $\hat{\mu}_{gt}$ satisfies $W_2(\mu_\varepsilon, \hat{\mu}_{gt}) \leq \varepsilon$. Moreover, if the latent codes are densely sampled from the latent manifold $\Omega$, we have $W_2(\mu,\mu_{gt})\leq 2\varepsilon,\ \mu$-almost surely.
\label{thm:thm1}
\end{theorem}
To avoid confusion, we omit the subscript $\varepsilon$ and denote $\mu_\varepsilon$ as $\mu$. With proof included in the supplementary material, this theorem tells us that as a continuous generalization of $\hat{\mu}_{gt}$, $\mu$ is a good approximation of $\mu_{gt}$. Also, we want to mention that $\tilde{T}$ is a piece-wise linear map that pushes forward $Uni([0,1]^d)$ to $\mu$, which makes the sampling from $\mu$ efficient and accurate.

\subsection{GAN Training from $\mu$}
\label{subsec:vanilla}
The GAN model computes the transport map from the continuous latent distribution $\mu$ to the data distribution on the manifold.
\[
\begin{CD}
(\mu, \Omega)@>{g_\xi}>> (\nu_{gt},\chi). 
\end{CD}
\]
Our GAN model is based on the vanilla GAN model proposed by Ian Goodfellow et.al \cite{Ian2014GAN}. The generator $g_\xi$ is used to generate new images by sampling from the latent distributin $\mu$, while the discriminator $d_\eta$ is used to discriminate if the distribution of the generated images are the same with that of the real images. The training process is formalized to be a min-max optimization problem:
\[
    \min_\xi \max_\eta \mathcal{L}(\xi,\eta),
\]
where the loss function is given by
\begin{equation}
    \mathcal{L}(\xi,\eta) = \mathcal{L}_{adv}+ \mathcal{L}_{feat} +\beta \mathcal{L}_{img}
    \label{eq: loss}
\end{equation}
In our model, the loss function consists of three terms, the image content loss $\mathcal{L}_{img}$, the feature loss $\mathcal{L}_{feat}$ and the adversarial loss $\mathcal{L}_{adv}$. Here $\beta>0$ is the weight of the content loss.

\textbf{Adversarial Loss}  We adopt the vanilla GAN model \cite{Ian2014GAN} based on the Kullback–Leibler (KL) divergence. The key difference between our model and the original GAN is that our latent samples are drawn from the data related latent distribution $\mu$, instead of a Gaussian distribution. The adversarial loss is given by:
\[
\begin{split}
    \mathcal{L}_{adv} &= \min_{\xi} \max_\zeta E_{x \sim \nu_{gt}}[log~ d_{\zeta}(x)] \\
    &+ E_{z\sim \mu}[log (1-d_{\zeta}(g_{\xi}(z)))]
    \label{eq: gan_loss}
\end{split}
\]
According to \cite{arjovsky2017wgan}, vanilla GAN is hard to converge because the supports of the distributions of real images and fake images may not intersect each other, which makes the KL divergence between them infinity. This issue is solved in our case, because (1) the training of AE gives a warm start to the generator, so at the beginning of the training, the generated distribution $g_{\xi\#}(\mu)$ is close to the real data distribution $\nu_{gt}$. (2) by delicate settings of the fake and real batches used to train the discriminator, we can keep the KL divergence between them converge well. In detail, as shown in Fig. \ref{fig:framework}, the fake batch is composed of both the reconstructed images from the real latent code (the orange circles) and the generated images from the generated latent code (the purple crosses), and the real batch includes both the real images corresponding to the real latent code and some randomly selected images.

\textbf{Content Loss}  Recall that the generator can produce two types of images: images reconstructed by real latent codes and images from generated latent codes.  Given a real sample $x_i$, its latent code is $z_i=f_\theta(x_i)$, the reconstructed image is $g_\xi(z_i)$. Each reconstructed image is represented as a triple $(x_i,z_i,g_\xi(z_i))$. Suppose there are $n$ reconstructed images in total, the content loss is given by
\begin{equation}
    \mathcal{L}_{img} = \frac{1}{n} \sum_{i=1}^n\|g_\xi(z_i) - x_i \|^2_2
\end{equation}
Where $g_\xi$ is the generator parameterized by $\xi$.

\textbf{Feature Loss}  We adopt the feature loss similar to that in \cite{Ledig2017SUR}. Given a reconstructed image triple $(x_i,z_i,g_\xi(z_i))$, we encode $g_\xi(z_i)$ by the encoder of AE. Ideally, the real image $x_i$ and the generated image $g_\xi(z_i)$ should be same, therefore their latent codes should be similar. We measure the difference between their latent codes by the feature loss. Furthermore, we can measure the difference between their intermediate features from different layers of the encoder.

Suppose the encoder is a network with $L$ layers, the output of the $l$th layer is denoted as $f_\theta^{(l)}$. The feature loss is given by
\[
    \mathcal{L}_{feat} := \frac{1}{n} \sum_{i=1}^n \sum_{l=1}^L \alpha^{(l)}\| f_\theta^{(l)}(x_i) - f_\theta^{(l)}\circ g_\xi(z_i)\|_2^2,
\]
Where $\alpha^{(l)}$ is the weight of the feature loss of the $l$-th layer. 

For reconstructed images $(x_i,z_i,g_\xi(z_i))$, the content loss and the feature loss force the generated image $g_\xi(z_i)$ to be the same with the real image $x_i$, therefore the manifold $g_\xi(\Omega)$ align well with the real data manifold $\chi$.

%% file: experiments.tex
\section{Expreiments}
\label{sec:exp}
To evaluate the proposed method, several experiments are conducted on simple dataset MNIST \cite{mnist2010} and complex datasets including Cifar10 \cite{cifar10}, CelebA \cite{celebA} and CelebA-HQ \cite{CelebAMask-HQ}. 

\begin{figure}[t]
\centering
\begin{tabular}{cc}
    \includegraphics[width=.5\linewidth]{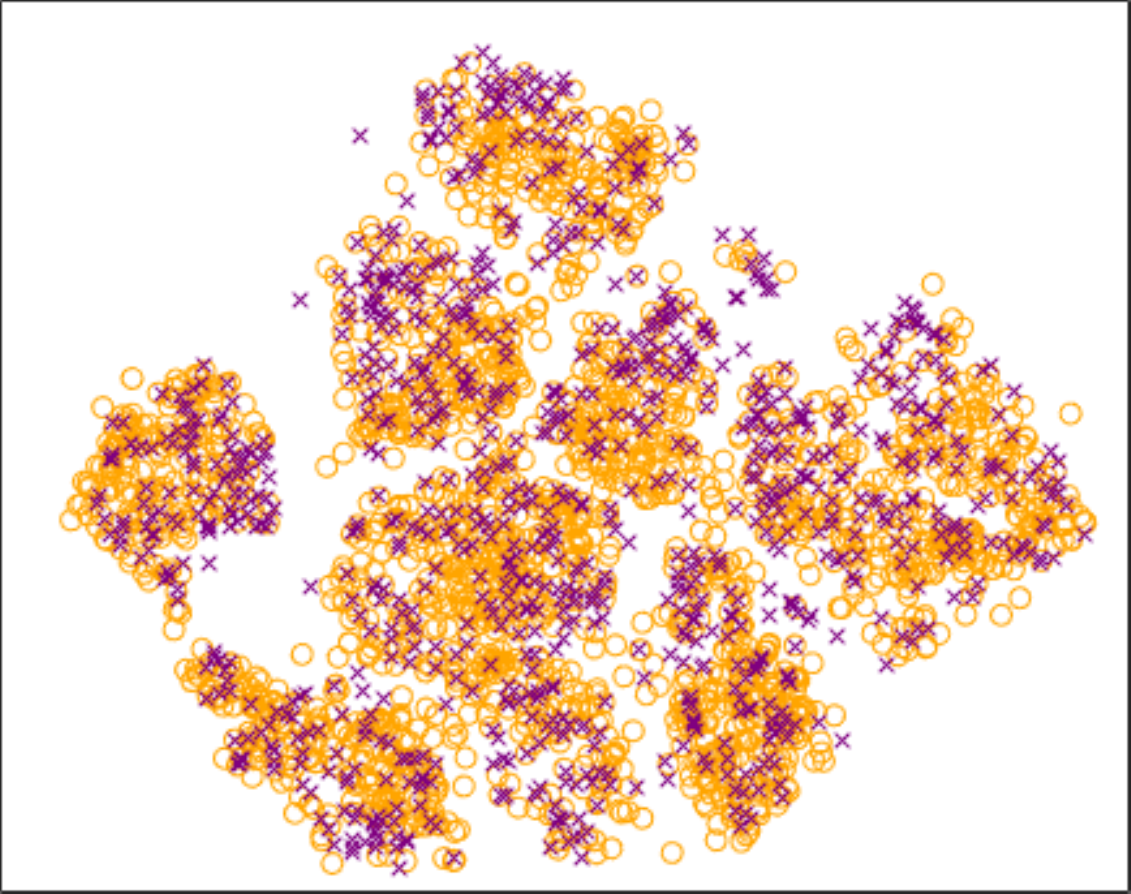} &
    \includegraphics[width=.4\linewidth]{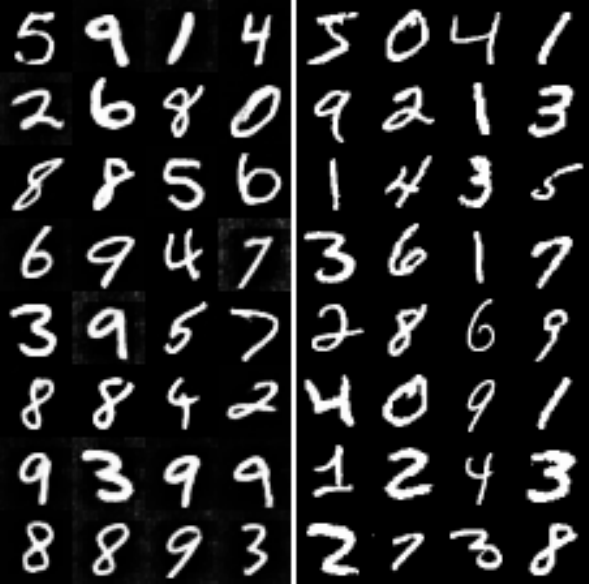}\\
    (a) & (b)
\end{tabular}
\caption{(a) Latent code distribution. The orange circles represent the fixed latent code and the purple crosses are the generated ones. (b) Comparison between the generated digits (left) and the real digits (right).}
    \label{fig:mnist}
    \vspace{-3mm}
\end{figure}

\begin{figure}[t]
\centering
\begin{tabular}{cc}
    \includegraphics[width=.45\linewidth]{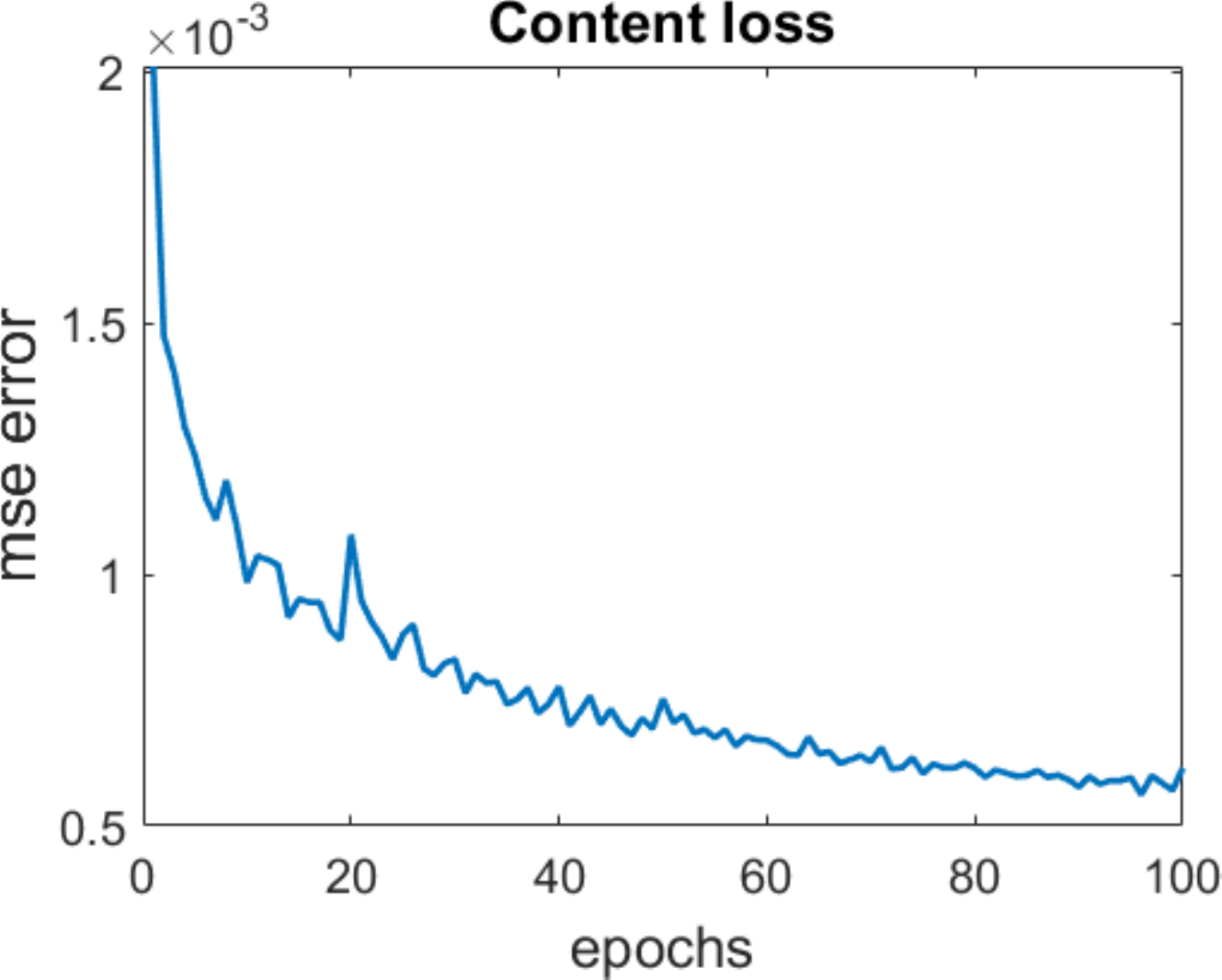} &
    \includegraphics[width=.45\linewidth]{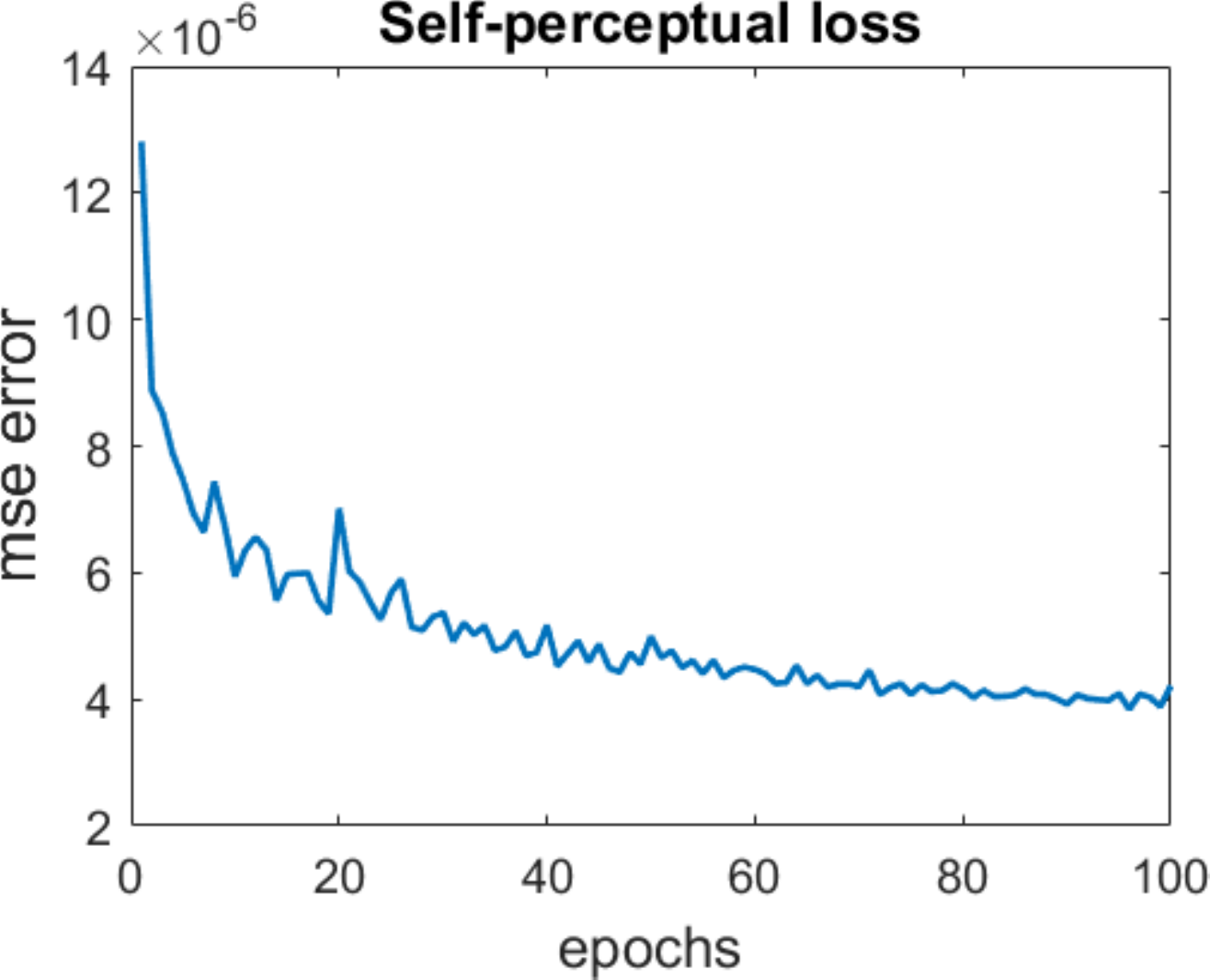}\\
    (a) & (b) \\
    \includegraphics[width=.45\linewidth]{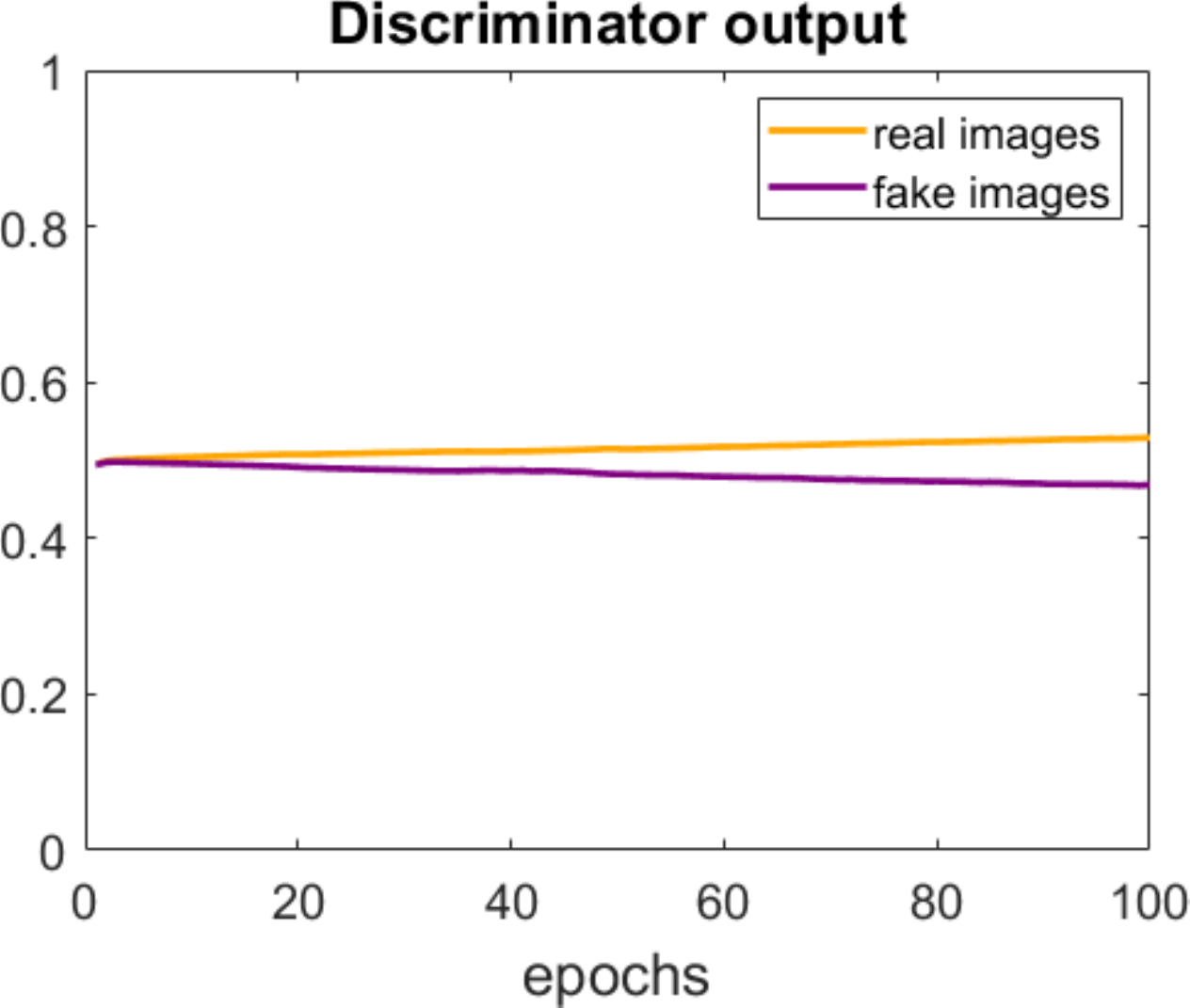} &
    \includegraphics[width=.45\linewidth]{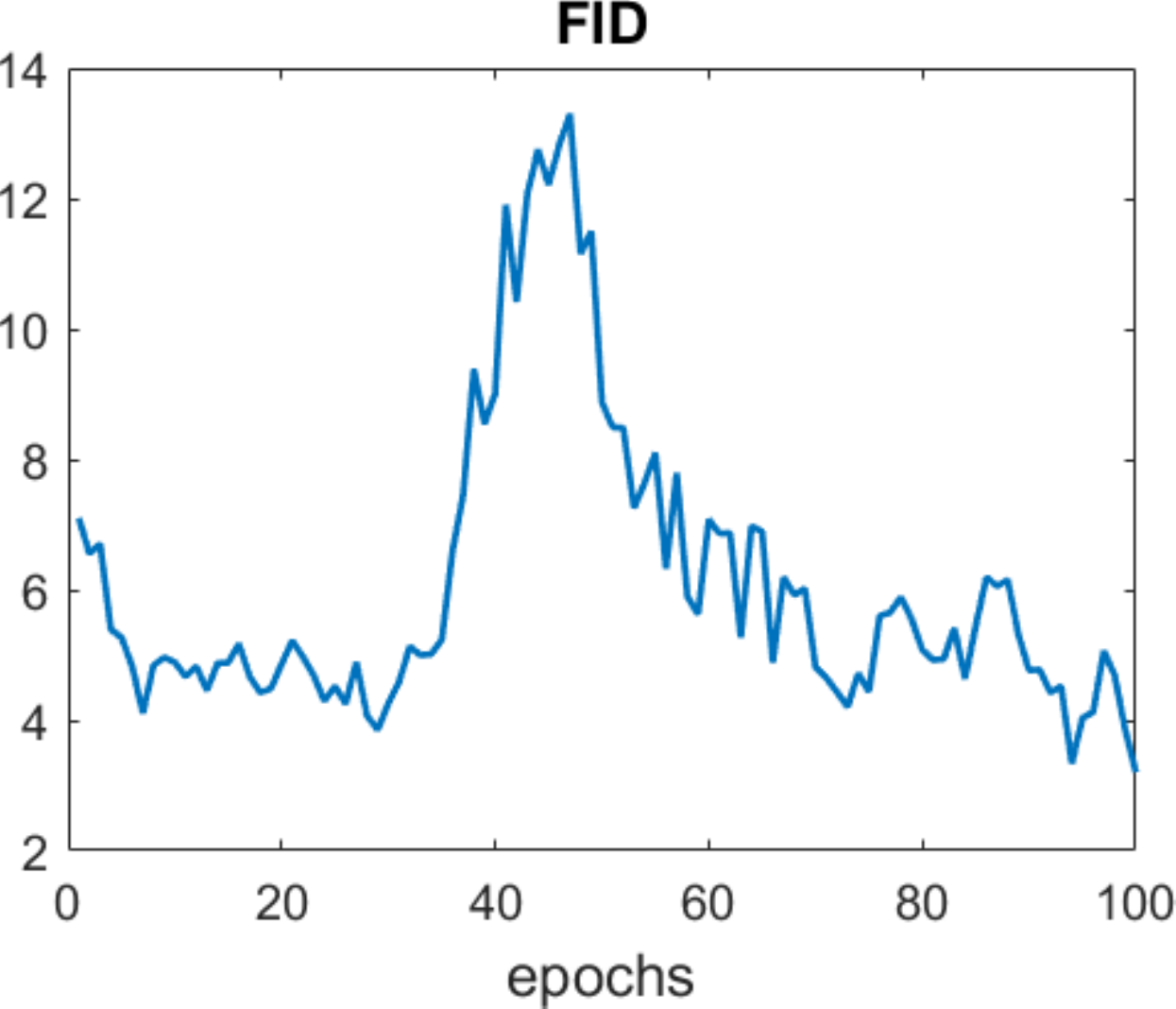}\\
    (c) & (d) 
\end{tabular}
\caption{The curves for training on MNIST dataset \cite{mnist2010} of each epoch, including the results of content loss (a) and self-perceptual loss (b), the discriminator output (c) and FIDs (d).}
    \label{fig:mnist_plot}
    \vspace{-5mm}
\end{figure}


\begin{figure*}[t]
\centering
\begin{tabularx}{\linewidth}{CCCCC}
    \includegraphics[width=\linewidth]{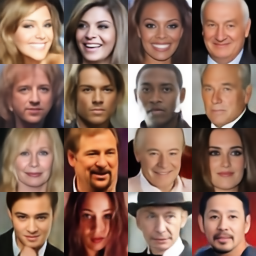} &
    \includegraphics[width=\linewidth]{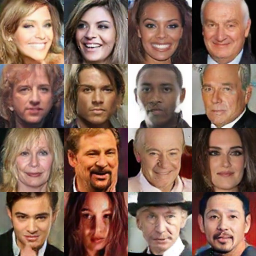} &
    \includegraphics[width=\linewidth]{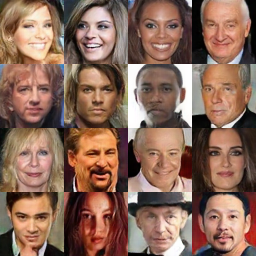} &
    \includegraphics[width=\linewidth]{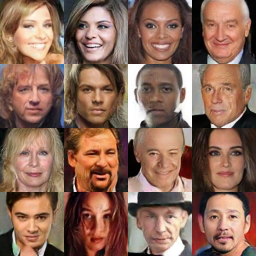} &
    \includegraphics[width=\linewidth]{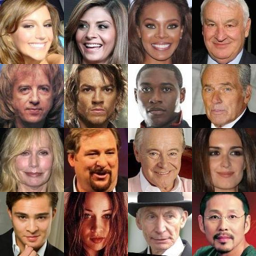} \\
    (a) Epoch 0 (AE-OT) & (b) Epoch 80 & (c) Epoch 160 & (d) Epoch 240 & (e) Ground-truth
\end{tabularx}
\caption{Evolution of the generator during training on the CelebA dataset \cite{celebA}. Reconstructed images from real latent codes at different epochs are shown.}
    \label{fig:evolution}
    \vspace{-3mm}
\end{figure*}

\begin{figure*}[t]
\centering
\begin{tabularx}{0.8\linewidth}{CCCC}
    \includegraphics[width=\linewidth]{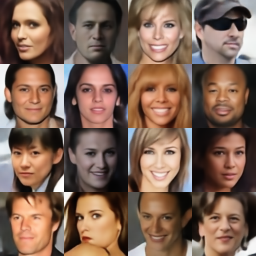} &
    \includegraphics[width=\linewidth]{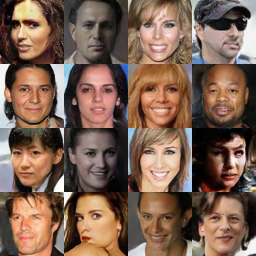} &
    \includegraphics[width=\linewidth]{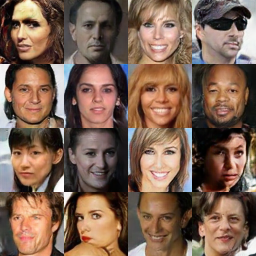} &
    \includegraphics[width=\linewidth]{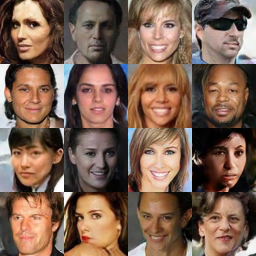} \\
    (a) Epoch 0 (AE-OT) & (b) Epoch 80 & (c) Epoch 160 & (d) Epoch 240
\end{tabularx}
\caption{Evolution of the generator during training on the CelebA dataset \cite{celebA}. Generated images from generated latent codes at different epochs are shown.}
    \label{fig:evolution_gen}
    \vspace{-3mm}
\end{figure*}

\textbf{Architecture} We adopt the InfoGAN \cite{infogan2016alec} architecture as our GAN model to train the MNIST dataset. The standard and ResNet models used to train the Cifar10 dataset are the same with those used by SNGAN \cite{miyato2018spectral}, and the architectures of WGAN-div \cite{wu2018wgan-div} are used to train the CelebA dataset. The framework of encoder is set to be the mirror of the generators/decoders. 

\textbf{Evaluation metrics}
To illustrate the performance of the proposed method, we adopt the commonly used Frechet Inception distance (FID) \cite{heusel2017fid} as our evaluation metrics. FID takes both the generated images and the real images into consideration. When the images are embedded into the feature space by inception network, two high dimensional Gaussian distributions are used to approximate the empirical distributions of the generated and real features, respectively. Finally, the FID is given by the difference between the two Gaussian distributions. Lower FID means better quality of the generated dataset. 
This metric has been proven to be effective in judging the performance of the generated models, and it serves as a standard for comparison with other works.

\textbf{Training details}
To get rid of the vanishing gradient problem and make the model converge better, we use the following three strategies:\\
\textit{(i) Train the discriminator using Batch Composition} There are two types of latent codes in our method: {\it the real latent codes} coming from encoding the real images by the encoder, and generated latent codes coming from the extended OT map. Correspondingly, there are two types of generated images, {\it the reconstructed images} from the real latent codes and {\it the generated images} from the generated latent codes.

To train the discriminator, both the fake batch and real batch are used. {\it The fake batch} consists of both randomly selected reconstructed images and generated images, and {\it the real batch} only includes real images, in which the first part has a one-to-one correspondence with the reconstructed images in the fake batch, as shown in Fig. \ref{fig:framework}. In all the experiments, the ratio between the number of generated images and reconstructed images in the fake batch is 3.

This strategy ensures that there is an overlap between the supports of the fake and real batches, so that the KL divergence is not infinity.

\textit{(ii) Different learning rate} For better training, we use different learning rates for the generator and the discriminator as suggested by Heusel et al. in \cite{heusel2017fid}. Specifically, we set the learning rate of the generator to be $lr_G = 2e-5$ and that of the discriminator to be $lr_D = lr_G / R$, where $R > 1$. This improves the stability of the training process.

\textit{(iii) Different inner steps} Another way to improve the training consistency of the whole framework is to set different update steps for the generator and discriminator. Namely, When the discriminator updated once, the generator updated $T$ times correspondingly.  This strategy is the opposite of training vanilla GANs, which typically require multiple discriminator update steps per generator update step. 

By setting $R$ and $T$, we can keep the discriminator output of the real images is slightly large than that of the generated images, which can better guide the training of the generator. For the MNIST dataset, $R=15$ and $T=3$; for the Cifar10 dataset, $R=25$ and $T=10$; and for the CelebA dataset, $R=15$ and $T=5$. In Eq. \ref{eq: loss}, $\beta=2000$ and $\alpha^{(l)}=0.06$ with $l<L$, where $L$ denotes the last layer of the encoder. $\alpha^{L}=2.0/\|Z\|_2$ is used to regularize the loss of the latent codes.

With the above settings and the warm initialization of the generator from the pre-trained decoder, for each dataset, the total epochs for training is set to be 500, which is far less than the training of GANs (usually 10k\textasciitilde 50k).

\begin{figure*}[t]
\centering
\begin{tabularx}{\linewidth}{CCCCC}
    \includegraphics[width=\linewidth]{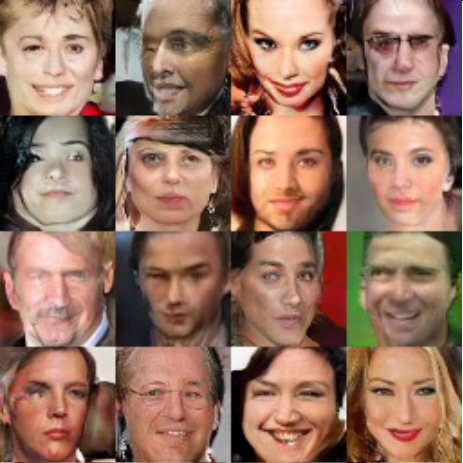} &
    \includegraphics[width=\linewidth]{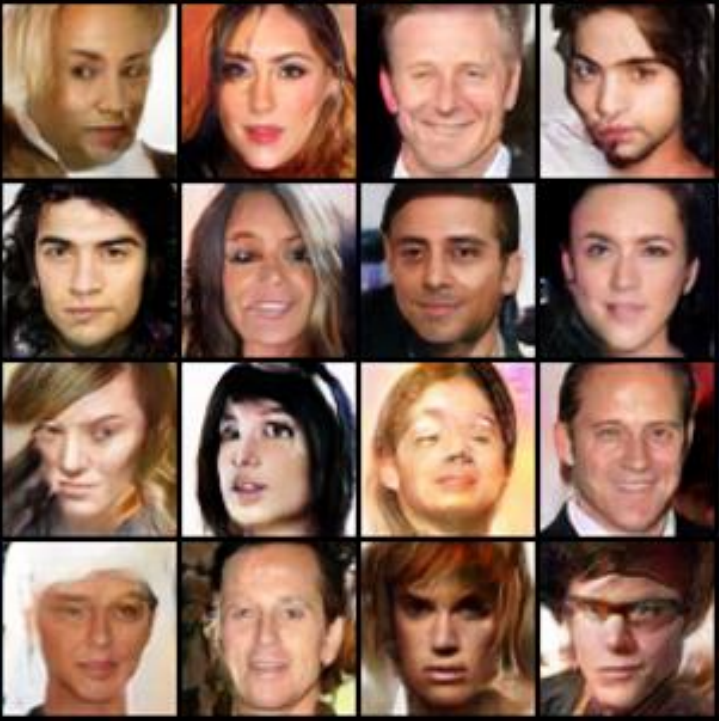} &
    \includegraphics[width=\linewidth]{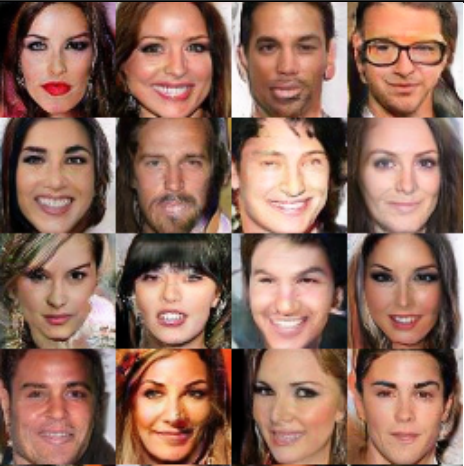} &
    \includegraphics[width=\linewidth]{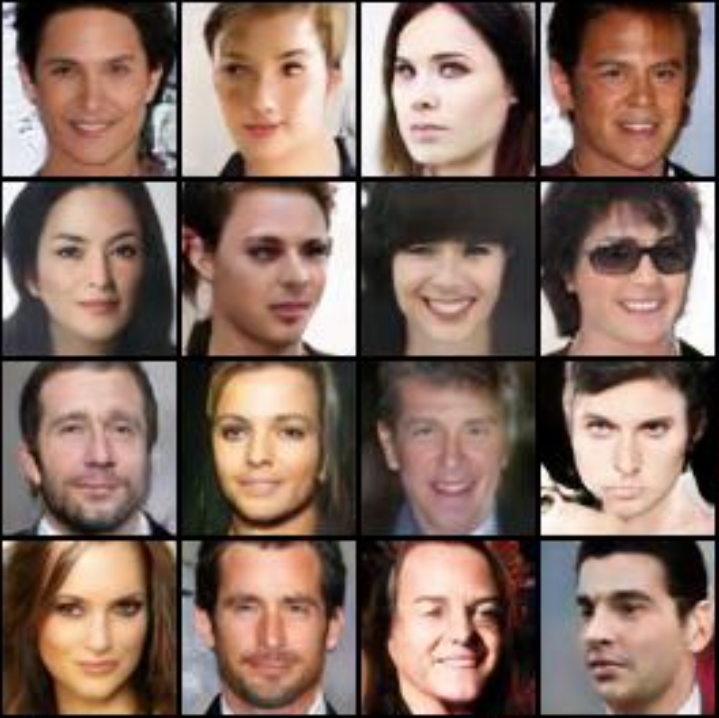} &
    \includegraphics[width=\linewidth]{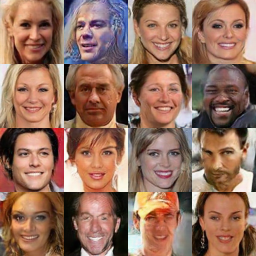} \\
    CT-GAN \cite{xu2019ctgan}  &  WGAN-GP \cite{gulrajani2017improvedgan} & WGAN-div \cite{wu2018wgan-div} & WGAN-QC \cite{liu2019wganqc} & Proposed method
\end{tabularx}
\caption{The visual comparison between the proposed method and the state-of-the-arts on CelebA dataset \cite{celebA} with ResNet architecture.}
    \label{fig:celebA_results}
    \vspace{-3mm}
\end{figure*}

\begin{figure*}[t]
\centering
\begin{tabularx}{\linewidth}{CCCCC}
    \includegraphics[width=\linewidth]{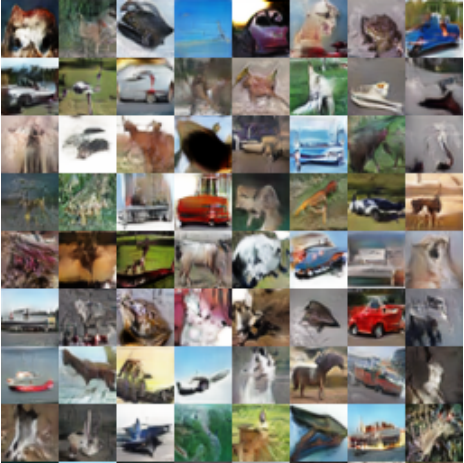} &
    \includegraphics[width=\linewidth]{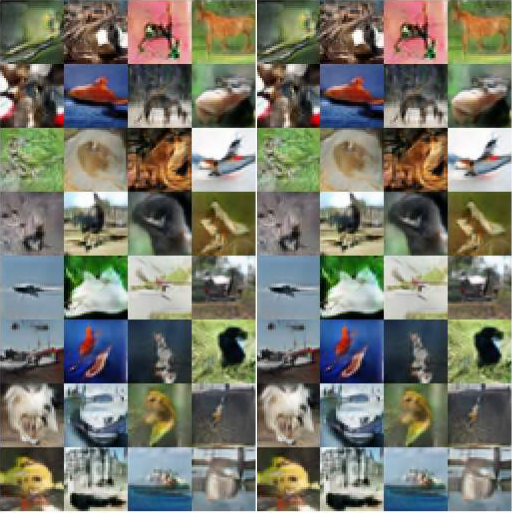} &
    \includegraphics[width=\linewidth]{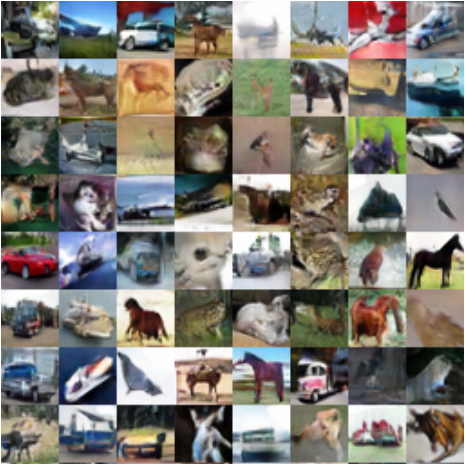} &
    \includegraphics[width=\linewidth]{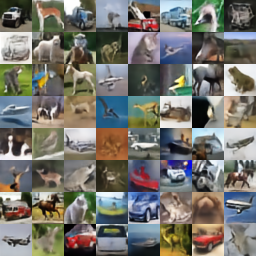} &
    \includegraphics[width=\linewidth]{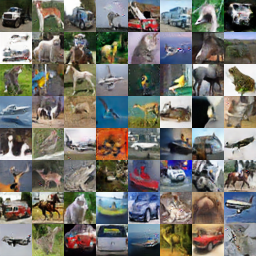} \\
    WGAN-GP \cite{gulrajani2017improvedgan} &  SNGAN \cite{miyato2018spectral}  & WGAN-div \cite{wu2018wgan-div} & AE-OT \cite{An2019AEOT} & Proposed method
\end{tabularx}
\caption{The visual comparison between the proposed method and the state-of-the-arts on Cifar10 dataset \cite{cifar10} with ResNet architecture.}
    \label{fig:cifar10_results}
    \vspace{-3mm}
\end{figure*}

\begin{figure*}[htbp!]
\centering
    \includegraphics[width=0.977\linewidth]{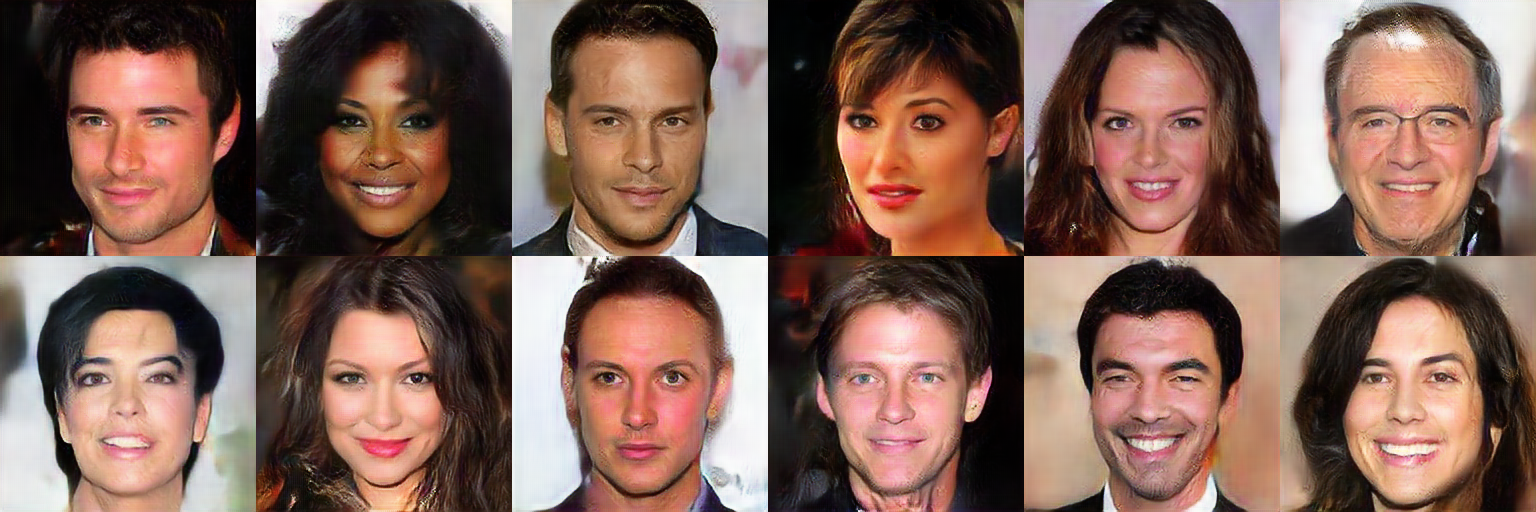} 
\caption{The generation results of CelebA-HQ by the proposed method.}
    \label{fig:celebA-HQ-gen}
    \vspace{-5mm}
\end{figure*}

\begin{table*}[htbp!]
\centering
\begin{tabular}{lcc c cc}
\hline
 & \multicolumn{2}{c}{CIFAR10} && \multicolumn{2}{c}{CelebA} \\
 \cline{2-3} \cline{5-6}
 & Standard & ResNet && Standard & ResNet \\
 \hline
WGAN-GP \cite{gulrajani2017improvedgan} & 40.2 & 19.6 && 21.2 & 18.4 \\
PGGAN \cite{karras2018pggan} & - & 18.8 && - & 16.3 \\
SNGAN \cite{miyato2018spectral} & 25.5 & 21.7 && - & - \\
WGAN-div \cite{wu2018wgan-div} & - & 18.1 && 17.5 & 15.2 \\
WGAN-QC \cite{liu2019wganqc} & - & - && - & 12.9 \\
AE-OT \cite{An2019AEOT} & 34.2 & 28.5 && 24.3 & 28.6 \\
\hline
AE-OT-GAN & 25.2 & 17.1 && 11.2 & 7.8\\
\hline
\end{tabular}
\caption{The comparison of FID between the proposed method and the state of the arts on Cifar10 and CelebA.}
\label{tab:fid}
\vspace{-5mm}
\end{table*}

\subsection{Convergence Analysis in MNIST}
In this experiment, we evaluate the performance of our proposed model on MNIST dataset \cite{mnist2010}, which can be well embedded into the $64$ dimensional latent space with the architecture of InfoGAN \cite{infogan2016alec}. In Fig. \ref{fig:mnist}(a), we visualize the real latent code (brown circles) and the generated latent codes (purple crosses) by t-SNE \cite{Maaten2008tsne}. It is obvious that the support of the real latent distribution and that of the generated distribution align well. Frame (b) of Fig.~\ref{fig:mnist} shows the comparison between the generated handwritten digits (left) and the real digits (right), which is very difficult for humans to distinguish.

To show the convergent property of the proposed method, we plot the related curves in Fig. \ref{fig:mnist_plot}. The frame (a) and (b) show the changes of the content loss about the images and latent codes, and both of them decrease monotonously. The frame (c) shows that the output of the discriminator for real images is only slightly larger than that for the fake images during the training process, which can help the generator generate more realistic digits. The frame (d) shows the evolution of FID and the final value is $3.2$. For MNIST dataset, the best known FIDs with the same InfoGAN architecture are $6.7$ and $6.4$, reported in \cite{lucic2018gans} and \cite{An2019AEOT} respectively. This shows our model outperforms state-of-the-art.

\subsection{Quality Evaluation on Complex Dataset}
In this section, we compare with the state-of-the-art methods quantitatively and qualitatively.

\textbf{Progressive Quality Improvement} Firstly, we show the evolution results of the proposed method in Fig. \ref{fig:evolution} and Fig. \ref{fig:evolution_gen} during GAN's training process. Quality of the generated images increases monotonously during the process. Images in first four frames of Fig.~\ref{fig:evolution} illustrates the results reconstructed from the real latent codes by the decoder, with the last frame showing the corresponding ground-truth input images. 
By examining the frames carefully, it is obvious that as the increase of the epochs, the generated images become sharper and sharper, and eventually they are very close to the ground truth. Fig.~\ref{fig:evolution_gen} shows the generated images from some generated latent codes (therefore, no corresponding real images). Similarly. the images become sharper as the increase of epochs. Here we need to state that the 0 epoch stage means the images are generated by the original decoder, which are equivalent to the outputs of an AE-OT model \cite{An2019AEOT}. Thus we can conclude that the proposed AE-OT-GAN improves the performance of AE-OT prominently.

\textbf{Comparison on CelebA and CIFAR 10} Secondly, we compared with the state-of-the-arts including WGAN-GP \cite{gulrajani2017improvedgan}, PGGAN \cite{karras2018pggan}, SNGAN \cite{miyato2018spectral}, CTGAN \cite{xu2019ctgan}, WGAN-div \cite{wu2018wgan-div}, WGAN-QC \cite{liu2019wganqc} and the recently proposed AE-OT model \cite{An2019AEOT} on Cifar10 \cite{cifar10} and CelebA \cite{celebA}. Tab. \ref{tab:fid} shows the FIDs of the our method and the comparisons trained under both the standard and ResNet architectures. The FID of other methods come from the listed papers except those of the AE-OT, which are directly computed by our model (the results of epoch 0). From the table we can see that our method gets much better results than others on the CelebA dataset, both under the standard and the ResNet architecture. Also, the generated faces of the proposed method have less flaws compared to other GANs, as shown on Fig. \ref{fig:celebA_results}. On Cifar10, the FIDs of our model are also comparable to the state-of-the-arts. And we also show some generated images on Fig. \ref{fig:cifar10_results}. The convergence curves for the both datasets can be found in the supplementary.

\begin{table}[htbp!]
\vspace{-2mm}
\small
\centering
\begin{tabular}{cccc}
\hline
PGGAN & WGAN-div & WGAN-QC & AE-OT-GAN \\
\hline
14.7  & 13.5     & 7.7     & 7.2  \\
\hline
\end{tabular}
\caption{The FIDs of the proposed method and the state-of-the-arts.}
\label{tab:fid-celeba-hq}
\vspace{-2mm}
\end{table}
\textbf{Experiment on CelebA-HQ} Furthermore, We also test the proposed method on images with high resolution, namely the CelebA-HQ dataset with image size to be 256x256. The architecture used to train the model is illustrated in the supplementary. The parameters in our model is far less than that of \cite{liu2019wganqc, wu2018wgan-div, karras2018pggan}, while the performance is better than theirs, as shown in Tab. \ref{tab:fid-celeba-hq}. We also display several images generated in Fig. \ref{fig:celebA-HQ-gen}, which are crisp and visually realistic.

%% file: conclusion.tex
\section{Conclusion and Future Work}
In this paper, we propose the AE-OT-GAN model which composes the AE-OT model and vanilla GAN together. By utilizing the merits of the both models, our method can generate high quality images without mode collapse nor mode mixture. Firstly, the images are embedded into the latent space by autoencoder, then the SDOT map from uniform distribution to the empirical distribution supported on the latent code is computed. Sampling from the latent distribution by applying the extended SDOT map, we can train our GAN model. Moreover, the paired latent code and images give us additional constraints about the generator. Using the FID as metric, we show that the proposed model is able to generate images comparable or better than the state of the arts. 